\pdfoutput=1

\documentclass[11pt]{article}

\usepackage[final]{acl}

\usepackage{times}
\usepackage{latexsym}

\usepackage{pifont}
\usepackage[hang]{footmisc}
\usepackage{graphicx}
\usepackage{dblfloatfix}
\usepackage{subcaption}
\usepackage{tablefootnote}
\usepackage{multicol}
\usepackage{multirow}
\usepackage{adjustbox}
\usepackage{comment}
\usepackage{placeins}
\usepackage{caption}
\usepackage{subcaption}
\usepackage{mathtools}
\usepackage{amsmath}
\usepackage{algorithm}
\usepackage{soul}
\usepackage{algpseudocode}
\algblock{Input}{EndInput}
\algnotext{EndInput}
\algblock{Parameters}{EndParameter}
\algnotext{EndParameter}

\usepackage[T1]{fontenc}

\usepackage[utf8]{inputenc}

\usepackage{microtype}

\usepackage{inconsolata}

\title{On The Persona-based Summarization of Domain-Specific Documents}

\author{$^1$Ankan Mullick \qquad $^1$Sombit Bose\thanks{\hspace{1mm} Authors contributed equally} \qquad $^1$Rounak Saha$^*$\qquad $^2$Ayan Kumar Bhowmick \qquad \\{\bf $^1$Pawan Goyal \qquad $^1$ Niloy Ganguly \qquad $^2$Prasenjit Dey \qquad $^2$Ravi Kokku} \\ \texttt{\{ankanm, sbcs.sombit, runk20\}@kgpian.iitkgp.ac.in}\\ \texttt{\{pawang, niloy\}@cse.iitkgp.ac.in}, \texttt{\{ayan, prasenjit, ravi\}@merlyn.org}\\ $^1$Computer Science and Engineering Department, IIT Kharagpur, India.  $^2$Emergence AI}

\begin{document}
\maketitle
\begin{abstract}

In an ever-expanding world of domain-specific knowledge, the increasing complexity of consuming, and storing information necessitates the generation of summaries from large information repositories. However, every persona of a domain has different requirements of information and hence their summarization. For example, in the healthcare domain, a persona-based (such as Doctor, Nurse, Patient etc.) approach is imperative to deliver targeted medical information efficiently. Persona-based summarization of domain-specific information by humans is a high cognitive load task and is generally not preferred. The summaries generated by two different humans have high variability and do not scale in cost and subject matter expertise as domains and personas grow. Further, AI-generated summaries using generic 
Large Language Models (LLMs) may not necessarily offer satisfactory accuracy for different domains unless they have been specifically trained on domain-specific data
and can also be very expensive to use in day-to-day operations. 
Our contribution in this paper is two-fold: 1) We present an approach to efficiently fine-tune a domain-specific small foundation LLM using a healthcare corpus and also show that we can effectively evaluate the summarization quality using AI-based critiquing. 2) We further show that AI-based critiquing has good concordance with Human-based critiquing of the summaries. Hence, such AI-based pipelines to generate domain-specific persona-based summaries can be easily scaled to other domains such as legal, enterprise documents, education etc. in a very efficient and cost-effective manner.

\end{abstract}

\section{Introduction}

In the rapidly expanding digital world, the exponential growth of domain-specific knowledge has posed unprecedented challenges in efficiently storing and consuming vast information repositories. With the increasing complexity of managing such information, the need for generation of precise and specific summaries becomes important. This need becomes particularly evident for domain-specific data as there exist different personas within a domain who have different information requirements which should be reflected in generated summaries. For instance, if we consider the healthcare domain, there exist diverse personas\footnote{\url{http://tiny.cc/x1guwz}} ranging from healthcare professionals like doctors and nurses to patients who require targeted information customized to their specific roles and comprehension levels.

Traditional generic approaches to summarization have often relied on humans to perform this high cognitive load task. However, as the volume and diversity of information burgeon with growing number of domains and personas, human generated persona-based summaries encounter limitations in scalability, cost-effectiveness, and consistency. The inherent subjectivity and variability among different human summarizers hinder the reliability and efficiency of such an approach. Although there have been several approaches in prior literature with focus on generic summaries through extractive and abstractive methods~\cite{paulus2017deep,erkan2004lexrank} as well as goal-oriented summaries~\cite{hayashi2021wikiasp,zhu2022transforming} but none of them have focused on persona-based summarization of domain-specific information. 
\newcite{goldsack2023overview, luo2022readability} focus on building layman-summarization comprehensible to non-technical audiences but do not differentiate the various technical summaries based on persona and they also do not use LLMs as an alternative evaluator. 
Our work differs in the sense that we develop a pipelined approach that generates persona-specific training summaries (doctor, patient, normal person), fine-tune small-size LLMs on this data, and use GPT-4 to efficiently evaluate summary quality.

One possible solution is to harness the power of generic large language models (LLMs) such as GPT-4 to automate the generation of persona-based summaries as such models have been used to generate data for other NLP tasks~\cite{sun2023evaluating,yu2023large}. 
ChatGPT~\footnote{https://chat.openai.com/} is also used in educational data generation~\cite{kieser2023educational, maddigan2023chat2vis}. However, AI-generated summaries using generic 
LLMs may not be guaranteed to achieve optimal accuracy across different domains unless they are trained on domain-specific data and they can also be very expensive to use for daily repeated inferences. In this paper, we take a step towards introducing a two-fold contribution aimed at overcoming these challenges of generating domain-specific persona-based summaries. Firstly, we present an efficient approach towards the training of domain-specific, small-sized Large Language Models (LLMs) on a corpus related to healthcare domain. 
Though data distillation from a stronger model for supervised fine-tuning is a standard method, the novelty of our work lies in the fact that we effectively employ the approach in this context to build a cost-optimised summarization framework catering to different healthcare persona given the scarcity of domain-specific data. This approach addresses the limitations of generic LLMs by aligning the trained model specifically to the intricacies of summaries in the healthcare domain. Moreover, we showcase the effectiveness of utilizing AI-based critiquing for the evaluation of summarization quality, providing a more automated and scalable solution.

Secondly, we demonstrate the strong agreement between AI-based and human-based critiquing of generated summaries, establishing the reliability of our proposed approach. This not only validates the effectiveness of domain-specific small LLM-based models in generating accurate summaries but also opens up avenues for scalability across diverse domains. The implications of our findings extend beyond healthcare, as the proposed AI pipeline can be seamlessly adapted to other domains, including legal, corporate documents, education, and more, offering a versatile and cost-effective solution for generating persona-based summaries.

\section{Proposed Framework}
We describe here our framework for training our small-size domain-specific LLM, generation of data for finetuning and evaluation, and other model baselines that we compare our model against.

\subsection{Dataset}

We create persona-based dataset (named `\textit{Persona-Data}') utilizing GPT-4\footnote{https://openai.com/gpt-4} with specific prompts on 1455 articles from the publicly available WebMD\footnote{https://www.webmd.com/} website. The mean ratio between summary length and document length is $0.2:1$. 
After data generation using GPT-4, we did a step by step validation of the generated summaries using an automated approach followed by manual verification to filter out the bad generations. Around $4.89\%$ of document-summary pairs were filtered out. We provide the detailed filtering steps with criteria and removal pairs (in \%) as shown in Table \ref{tab:filtering}. 

\begin{table}[t]
\centering
\begin{adjustbox}{width=\linewidth}
\begin{tabular}{|l|c|}
\hline
\textbf{Filtering Steps with Criteria (removed)} & \textbf{\%} \\ \hline
Step 1: Too many special characters and other string (HTML tags and \# )	& 1.52  \\ \hline
Step 2: Incomplete Summary (By checking punctuations)  & 0.86  \\ \hline
Step 3: Conflict identification - Very similar summaries of different persona	& 1.12   \\ \hline
\begin{tabular}{@{}l@{}} Step 4: If the summary contains Medical Terms or numbers not present in the document (using QuickUMLS - \\ \url{https://github.com/Georgetown-IR-Lab/QuickUMLS/})  \\ \end{tabular} & 1.39 \\\hline
Overall summaries filtered out	& 4.89   \\ \hline
\end{tabular}
\end{adjustbox}
\caption{Step-by-Step Data Filtering}
\label{tab:filtering}
\vspace{-6mm}
\end{table}


These articles, related to healthcare, form the basis for creation of their summaries related to three distinct persona: 
\textit{(a) Doctor:} Summaries focus on medical terminology, guidelines and provide detailed technical information suitable for medical professionals.
\textit{(b) Patient:} Summaries are easily understandable, addressing patient concerns without excessive technical jargon, focusing on top-level information.
\textit{(c) Normal Person:} Summaries are tailored for a general audience without medical background, presented in simple language and engaging for laypersons while avoiding technical terms. The dataset comprises 1091, 73 and 291 articles for training, validation, and testing respectively.  
Additionally, we select 50 WebMD articles and generate manual summaries for three personas (termed as \textit{Annotated-Data}) using the Prolific\footnote{https://www.prolific.com/} annotation platform and Doctors to evaluate the GPT-4 generated summaries against human curated summaries\footnote{Code/Dataset details are in \url{https://github.com/ankan2/persona-healthcare}}
(Details are in Appendix \ref{appendix : human_annotations}).

\begin{table*}[!ht]
    \centering
    \vspace{-6mm}
    \begin{adjustbox}{width=0.81\linewidth}
    \begin{tabular}{|c|c|c|c|c|c|c|c|c|c|}
        \hline
        \textbf{Model} & \textbf{Rouge1} & \textbf{Rouge2} & \textbf{RougeL}& \textbf{Meteor} & \textbf{Bleu} & \textbf{BERT-Prec} & \textbf{BERT-Rec} & \textbf{BERT-F1} \\
        \hline
        Falcon & 23.3 & 5.0 & 12.8 & 12.4 & 1.3 & 78.6 & 73.2 & 75.8\\
        \hline
        BART & 23.6 & 8.1 & 15.0 & 11.3 & 0.8 & 83.8 & 74.9 & 79.1\\
        \hline
        Pegasus & 27.8 & 8.3 & 16.2 & 14.3 & 2.0 & 80.7 & 75.0 & 77.8\\
        \hline
        T5-FT & 42.6 & 15.6 &  23.7 & 30.2 & 11.4 & 84.2 & 80.7 & 82.4\\
        \hline
        FT5-FT & 41.2 & 15.5 &  23.3 & 29.0 & 11.2 & 83.9 & 79.9 & 81.9\\
        \hline
        LED-B & 33.3 & 9.3 & 17.4  & 17.3 & 3.3 & 80.7 & 76.7 & 78.6\\
        \hline
        LED-L & 39.8 & 12.3 & 25.6 & 25.3 & 9.3 & 82.5 & 78.7 & 80.6\\
        \hline
        L-V-7b & 14.2 & 4.0 & 8.5 & 6.7 & 0.6 & 72.0 & 65.7 & 68.7\\
        \hline
        L-V-13b & 24.1 & 7.3 &  14.3 & 11.5 & 1.1 & 80.8 & 73.5 & 77.0\\
        \hline
        L-V-70b & 25.1 & 7.2 & 13.7 & 16.2 & 3.7 & 65.7 & 64.6 & 65.1\\
        \hline
        L-F-7b & 45.9 & 17.2 & 25.2 & 32.6 & 12.5 & 83.6 & 82.6 & 83.1\\
        \hline
        \textbf{L-F-13b} & \textbf{53.7} & \textbf{24.3} & \textbf{33.8} & \textbf{38.3} & \textbf{18.4} &  \textbf{87.4} & \textbf{85.3} & \textbf{86.3}\\ 
        \hline
    \end{tabular}
    \end{adjustbox}
    \caption{Traditional Metrics' based evaluation results (all are in \%)}
    \label{tab:webmd-rouge}
    \vspace{-3mm}
\end{table*}

\begin{table*}[b]
    \centering
    \vspace{-1mm}
    \begin{adjustbox}{width=0.79\linewidth}
    \begin{tabular}{|c|c|c|c|c|c|c|c|c|c|}
        \hline
        \textbf{Model} & \textbf{Rouge1} & \textbf{Rouge2} & \textbf{RougeL}& \textbf{Meteor} & \textbf{Bleu} & \textbf{BERT-Prec} & \textbf{BERT-Rec} & \textbf{BERT-F1} \\
        \hline
        Doctor	& 53.9	& 24.7	& 34.0	& 37.0	& 18.7	& 88.0	& 85.1	& 86.5\\
        \hline
        Patient	& 53.5	& 24.2	& 33.5	& 36.7	& 18.3	& 87.2	& 84.8	& 86.0\\\hline
Nor-Per	& 53.6	& 23.9	& 33.9	& 41.0	& 18.1	& 86.9	& 86.1	& 86.5\\\hline
\textbf{Average} & \textbf{53.7} & \textbf{24.3} & \textbf{33.8} & \textbf{38.3} & \textbf{18.4} & \textbf{87.4} & \textbf{85.3} & \textbf{86.3}\\ 
        \hline
    \end{tabular}
    \end{adjustbox}
    \vspace{-2mm}
    \caption{Traditional Metrics' based evaluation on different persona using Llama2-13b Finetuning model (in \%)}
    \label{tab:webmd-persona-rouge}
    \vspace{-2mm}
\end{table*}

\subsection{Model Architecture}
Our training process consists of employing small foundation LLMs such as Llama2 and finetuning such models on the training set of the WebMD data described above. Specifically, we use supervised fine-tuning approach~\cite{li2023label} on the pre-trained vanilla model versions of Llama2. 

\noindent \textbf{Llama2}\footnote{https://ai.meta.com/llama/}: 
We perform supervised fine-tuning (SFT) on the pretrained vanilla Llama2-7b and Llama2-13b models using the training data that consists of prompt-completion pairs where the prompt comprises a WebMD article in the training set and the instruction to generate a summary based on a persona (doctor/patient/normal person) and the completion is the corresponding persona-based summary generated using GPT-4. We use a parameter efficient finetuning approach i.e. Quantized Low-Rank Adaptation (QLoRA)~\cite{dettmers2023qlora} to optimize the training process. After training, the finetuned Llama2-7b and Llama2-13b models (referred to as L-F-7b and L-F-13b respectively) acquire the ability to generate a persona-based summary for a given medical article depending on the persona specified in the prompt.   

\noindent \textbf{Baselines:} For comparison with our finetuned models on the persona-based summary generation task, we use different state-of-the-art models as baselines such as Falcon 7b-instruction tuned model~\cite{penedo2023refinedweb}, BART-large~\cite{lewis2019bart}, instruction-tuned Pegasus~\cite{zhang2020pegasus} and Longformer~\cite{beltagy2020longformer} Base (LED-B), Large (LED-L)\footnote{https://huggingface.co/allenai/led-base-16384}. Besides these, we also use finetuned versions of T5-Large~\cite{raffel2020exploring} (T5-FT) and Flan-T5-Large~\cite{chung2022scaling} (FT5-FT) on our training data as baselines. Further, we also compare the performance with the different vanilla Llama2 model variants (7b, 13b, 70b referred to as L-V-7b, L-V-13b and L-V-70b respectively).

\section{Evaluation and Results}
We evaluate the performance of our finetuned models (L-F-7b and L-F-13b) in terms of generating high quality persona-based summaries for medical articles and compare against the baseline models. 

\noindent \textbf{Evaluation metrics:} Our evaluation relies on two different approaches: \\\textit{(i) Traditional} - Here we use traditional metrics such as Rouge [1, 2 and L]~\cite{lin2004rouge}, Meteor~\cite{banerjee2005meteor}, Bleu~\cite{papineni2002bleu},  BERTScore~\cite{zhang2019bertscore} [Precision (BERT-Prec), Recall (BERT-Rec) and F1-score (BERT-F1)] to assess the quality of generated summaries. \\
\textit{(ii) GPT-4 critique} - Here we use the GPT-4 LLM as a critic to evaluate the quality of the model generated summaries against the gold standard GPT-4 generated summary (Section 2) from different dimensions. Specifically, we provide suitable critique based prompts to GPT-4 where we evaluate the summaries based on a set of five predefined criterias (termed as \textit{GPT-4 criteria}) defined below:\\
\textit{Criteria 1: Relevance (Rel):} The extent to which the generated persona-based summary is relevant to the intended persona (doctor/patient/normal person) given the document.\\
\textit{Criteria 2: Coverage (Cov):} The extent to which the generated persona-based summary correctly covers important key points described in the gold standard persona-based summary of the document.\\
\textit{Criteria 3: Impurity (Imp):} The extent to which the persona-based summary does not contain information specific to all other possible personas $\{persona\_set$ - persona$\}$. \\
\textit{Criteria 4: Quality (Qlt):} The extent to which the persona-based summary is of overall good quality from the perspective of the intended persona. \\
\textit{Criteria 5: Goodness (Gds):} Extending from 4, we manually verify the goodness of the summary.\\
(Details of these criteria with prompts are provided in Appendix \ref{appendix : prompts}).

\noindent \textbf{Results:} We provide a comparison of our finetuned models against the baselines in terms of both traditional metrics as well as GPT-4 criteria in Tables~\ref{tab:webmd-rouge} and~\ref{tab:webmd-gpt4-critic} respectively (all values are in $\%$) on the WebMD test set of size $873$ (prompt specific to three different persona each for $291$ articles). Table~\ref{tab:webmd-rouge} infers that both our finetuned models (L-F-7b and L-F-13b) achieve superior performance compared to the baseline methods in terms of traditional metrics. In fact, finetuned Llama2-13b (L-F-13b) outperforms the baselines in terms of all the traditional metrics, demonstrating the superiority of our finetuning approach which helps to adapt the model to the healthcare domain and perform better on specific applications such as persona-based summarization. Similar observation holds true when we compare the values of the GPT-4 critique based criteria shown in Table \ref{tab:webmd-gpt4-critic}. Here we also compare the quality of the gold standard GPT-4 generated summaries in terms of the GPT-4 critique based criteria. We find that the finetuned Llama2-13b model (L-F-13b) can generate summaries pretty close in quality to the gold standard, while being much faster in terms of training and inference time as well as cost-effective and cheaper in terms of memory requirement.

\begin{table}[!ht]
    \centering
    \vspace{-2mm}
    \begin{adjustbox}{width=0.78\linewidth}
    \begin{tabular}{|c|c|c|c|c|c|}
        \hline
        \textbf{Model} & \textbf{Rel} & \textbf{Cov} & \textbf{Imp} & \textbf{Qlt} & \textbf{Gds}\\\hline
        Falcon & 56.3 & 45.4 & 82.0 & 50.6 & 46.8\\
        \hline
        BART & 65.4 & 42.6 & 84.8 & 49.7 & 25.8\\
        \hline
        Pegasus& 47.7 & 33.0 & 74.0 & 36.1 & 11.5\\
        \hline
        T5-FT & 72.4 & 70.1 & 84.9 & 67.6 & 78.1\\
        \hline
        FT5-FT & 72.2 & 70.2 & 88.0 & 68.3 & 80.3\\
        \hline
        LED-B & 36.1 & 19.2 & 79.3 & 31.3 & 17.6\\
        \hline
        LED-L & 69.1 & 56.2 & 82.3 & 59.3 & 56.6 \\\hline
        L-V-7b & 19.1 & 18.4 & 41.3 & 16.6 & 15.2\\
        \hline
        L-V-13b & 32.1 & 29.1 & 73.1 & 28.5 & 23.4\\
        \hline
        L-V-70b & 49.7 & 45.1 & 78.0 & 46.4 & 47.8\\
        \hline
        L-F-7b & 75.8 & 58.7 & 85.8 & 63.8 & 58.6\\
        \hline
        \textbf{L-F-13b} & \textbf{93.5} & \textbf{90.1} & \textbf{91.7} & \textbf{88.5} & \textbf{99.1}\\\hline
        GPT-4 & 98.2 & 96.3 & 98.6 & 98.5 & 99.7\\
        \hline
    \end{tabular}
    \end{adjustbox}
        \vspace{-2mm}
    \caption{GPT-4 Critique evaluation results (in \%)}
    \label{tab:webmd-gpt4-critic}
        \vspace{-4mm}
\end{table}

\noindent \textbf{Framework:} We use 80GB A100 GPU, 210MHz clock cycle along with NLTK/SpaCy python packages for all experiments. For 6 epochs, Llama2-13b takes 20 hrs for finetuning and 3 hrs for inference, Llama2-7b takes 8 hrs for finetuning and 2.5 hrs for inference (Details in Appendix \ref{time-gpu}).

\noindent \textbf{Ablation study:} Here we investigate the quality of persona-based summaries generated by different variations of our best performing finetuned Llama2-13b (L-F-13b) model on WebMD test set:

\noindent \textbf{(A) Performance specific to different persona:} 
Table \ref{tab:webmd-persona-rouge} and \ref{tab:webmd-3persona-critic} shows the 
performance in terms of standard evaluation metrics and outcomes of GPT-4 critique based criteria for each of the three persona [Doctor, Patient and Normal Person (Nor-Per)] for the best performing Llama2-13b Finetuned model (L-F-13b). We observe that the model performs uniformly across the three persona which confirms that our finetuned model generalizes well across multiple persona, generating distinct persona-based summaries for the same medical article. 

\begin{table}[!ht]
    \centering
    \begin{adjustbox}{width=0.78\linewidth}
    \begin{tabular}{|c|c|c|c|c|c|}
        \hline
        \textbf{Persona} & \textbf{Rel} & \textbf{Cov} & \textbf{Imp} & \textbf{Qlt} & \textbf{Gds}\\
        \hline
        Doctor & 90.0 & 89.1 & 91.0 & 86.2 & 98.8\\
        \hline
        Patient & 94.4 & 90.4 & 92.2 & 88.5 & 99.0\\
        \hline
        Nor-Per & 93.2 & 91.0 & 91.8 & 87.7 & 99.3\\
        \hline
        Average & 93.5 & 90.1 & 91.7 & 88.5 & 99.1\\
        \hline
    \end{tabular}
    \end{adjustbox}
        \vspace{-2mm}
    \caption{GPT-4 critique on different persona using \\ Llama2-13b Finetuning model (in \%)}
    \label{tab:webmd-3persona-critic}
        \vspace{-3mm}
\end{table}

\noindent \textbf{(B) Validation of GPT-4 generated gold standard summaries:} To verify the robustness of the GPT-4 generated summaries for the WebMD articles 
and to mitigate the GPT-4 introduced inherent bias in generated summaries, we perform different types of human annotation experiments: 

\textbf{(i) Persona-based Summary of GPT-4:} We randomly select $50$ different WebMD articles and provide three different persona-based (doctor, patient, normal person) summaries (without their actual labels) to three different doctors with domain
knowledge expertise along with a good working proficiency in English and ask them to identify 
the intended persona, i.e. - which summary belongs to which specific persona. Initial human labeling is done by
two doctors and any annotation discrepancy is
checked and resolved by the third doctor after discussing with others. The inter-annotator agreement 
is found to be $0.91$.
On comparing with actual persona labels, we found that human labels have $86.67\%$ accuracy for correctly identifying the actual persona which shows the reliability of GPT-4 generated persona-based summaries. 

\textbf{(ii) Content Quality Check:} We ask human annotators (doctors) to annotate summaries on the basis of whether the persona-based summary is relevant while correctly covering appropriate key points based on information need of different persona and its overall usefulness. 96\% of the GPT-4 generated summaries for different personas are found to be useful by human annotators (doctors).

\textbf{(iii) GPT-4 Generated and Ground Truth Summary Check:} 
Both doctors with domain expertise and GPT-4 evaluate 50 document-summary pairs in terms of whether the persona-based summary is relevant and correctly covers appropriate key points based on information need of different persona, each for GPT-4 generated summaries and ground truth summaries generated by annotators. We obtain an inter-annotator agreement of $0.893$ which signifies strong consensus between human and GPT-4 based evaluation. We separately test our best fine-tuned (Llama2-13b-FT) model on the human-generated summaries (50 articles) and obtain the following scores for different metrics - Traditional: R1-52.9, R2-24.1, RL-33.2, Meteor-38.7, Bleu-18.0, Bert-P-87.7, Bert-R-84.9, Bert-F1-86.3; GPT-4 criteria: Rel-91.2, Cov-90.8, Imp-90.4, Qlt-88.7, Gds-98.5. This shows that there is strong alignment between results of on human generated and GPT-4 generated summaries, signifying the high quality of GPT-4 generated summaries.

\noindent \textbf{(C) Validation of finetuned model generated summaries:} To further investigate the reliability of our finetuned model generated summaries, we choose $50$ different WebMD articles and provide 
persona-based summaries for each persona (generated by GPT-4 in ground truth v/s Llama2-13b finetuned model generated) to two doctors to annotate: (i) whether finetuned generated summary is better, (ii) Both are Good, (iii) Ground Truth/GPT-4 summary is better and (iv) Both are bad. We find that for - 20\% cases Llama2-13b finetuned model summaries are better (i), for 50\% cases both finetuned and ground truth generated summaries are good (ii) and rest 30\% cases ground truth generated summaries are better (iii) and no instance is found where both performs bad (iv).

\noindent \textbf{(D) Different LLM Evaluators:} We evaluate the fine-tuned model generated summaries in the test set with Gemini model\footnote{\url{https://gemini.google.com/}} keeping the same prompts and criteria as used earlier for GPT-4 and the obtain values of the same LLM based criteria are: Rel - 95.2, Cov - 92.4, Imp - 87.6, Qlt - 90.7, Gds - 99.4. Thus, Gemini scores are also aligned with GPT-4 scores with a correlation coefficient of 0.808 (Gemini provides higher scores for all criteria except Criteria 3 - `Imp'). This verifies that the GPT-4 based evaluation is impartial and robust.

\noindent \textbf{(E) Llama2-13b performance on Other data:} We test our best performing Llama2-13b finetuned model on healthcare domain articles of OASUM~\cite{yang2022oasum} dataset which is publicly available. We select OASUM articles with aspects related to healthcare [Death, Diagnosis, Differential Diagnosis and Diagnosis Classification] and obtain $234$ such documents. We perform GPT-critique based evaluation and observe that $82.77\%$ of the summaries are labeled as good which signifies the robustness of our model in terms of generating high quality summaries. 

\section{Conclusion}

In this paper, we propose a framework for the efficient training of a small foundation LLM on AI-generated datasets to obtain high quality domain-specific persona-based summaries. Our focus is on training a finetuned version of Llama2 on a corpus related to healthcare domain such that the trained model captures the intricacies of persona-based summaries in healthcare domain. We also demonstrate the effectiveness of using AI-based critiquing for the evaluation of the model generated summaries, providing a more automated and scalable solution. Our experiments also reveal the superior quality of persona-based summaries generated by our finetuned model compared to contemporary  baselines.
Further, AI-based critiquing of the summaries show high inter-annotator agreement with human-based critiquing methods, further confirming the effectiveness of our proposed approach. We plan to extend our work in generating accurate persona-based summaries for documents in other domains such as legal, enterprises, education and more, which is the focus of our future work.

\section*{Acknowledgements}
The work was supported in part by a research grant from IIT KHARAGPUR AI4ICPS I HUB FOUNDATION.

\section*{Limitation and Discussion}

There are a few limitations in our works - (i) Not all LLMs are useful (similar to GPT-4) to generate personalized contents properly - like GPT2, GPT3.5, Llama2-vanilla models do not perform very well mostly due to hallucination and not covering important informations. (ii) We only explore the data from healthcare domain, but we plan to extend our work to other domains such as legal, corporate and education among others. (iii) Our experimental dataset is only English in heathcare domain, we wish to extend the work in multilingual setup, specifically for low-resource settings in diverse domains. (iv) Prompt command is very important. Unless, we specifically mention with explanation in Prompt about different {\textit{Persona}} (Doctor/Patient/Normal Person), GPT-4 does not perform well to generate appropriate summary. 

\section*{Ethical Concerns}
We use the publicly available content of the WebMD platform for non-commercial and academic purpose only without violating any ethical concerns. The dataset neither
reveals any personal sensitive information of the
patients nor any toxic statement. Consent has been taken from all annotators including doctors. For experiments, we use publicly available free frameworks - Llama2, Falcon, BART, Pegasus, T5-FT, Flan-T5 (FT5-FT), LED-Base, LED-Large, Llama2-7b, 13b and 70b - vanilla and finetune. 

\bibliography{custom}

\begin{thebibliography}{53}
\expandafter\ifx\csname natexlab\endcsname\relax\def\natexlab#1{#1}\fi

\bibitem[{Akhtar et~al.(2017)Akhtar, Zubair, Kumar, and Ahmad}]{akhtar2017aspect}
Nadeem Akhtar, Nashez Zubair, Abhishek Kumar, and Tameem Ahmad. 2017.
\newblock Aspect based sentiment oriented summarization of hotel reviews.
\newblock \emph{Procedia computer science}, 115:563--571.

\bibitem[{Banerjee and Lavie(2005)}]{banerjee2005meteor}
Satanjeev Banerjee and Alon Lavie. 2005.
\newblock Meteor: An automatic metric for mt evaluation with improved correlation with human judgments.
\newblock In \emph{Proceedings of the acl workshop on intrinsic and extrinsic evaluation measures for machine translation and/or summarization}, pages 65--72.

\bibitem[{Beltagy et~al.(2020)Beltagy, Peters, and Cohan}]{beltagy2020longformer}
Iz~Beltagy, Matthew~E Peters, and Arman Cohan. 2020.
\newblock Longformer: The long-document transformer.
\newblock \emph{arXiv preprint arXiv:2004.05150}.

\bibitem[{Chan et~al.(2023)Chan, Chen, Su, Yu, Xue, Zhang, Fu, and Liu}]{chan2023chateval}
Chi-Min Chan, Weize Chen, Yusheng Su, Jianxuan Yu, Wei Xue, Shanghang Zhang, Jie Fu, and Zhiyuan Liu. 2023.
\newblock Chateval: Towards better llm-based evaluators through multi-agent debate.
\newblock \emph{arXiv preprint arXiv:2308.07201}.

\bibitem[{Chintagunta et~al.(2021)Chintagunta, Katariya, Amatriain, and Kannan}]{chintagunta2021medically}
Bharath Chintagunta, Namit Katariya, Xavier Amatriain, and Anitha Kannan. 2021.
\newblock Medically aware gpt-3 as a data generator for medical dialogue summarization.
\newblock In \emph{Machine Learning for Healthcare Conference}, pages 354--372. PMLR.

\bibitem[{Chopra et~al.(2016)Chopra, Auli, and Rush}]{chopra2016abstractive}
Sumit Chopra, Michael Auli, and Alexander~M Rush. 2016.
\newblock Abstractive sentence summarization with attentive recurrent neural networks.
\newblock In \emph{Proceedings of the 2016 conference of the North American chapter of the association for computational linguistics: human language technologies}, pages 93--98.

\bibitem[{Chung et~al.(2022)Chung, Hou, Longpre, Zoph, Tay, Fedus, Li, Wang, Dehghani, Brahma et~al.}]{chung2022scaling}
Hyung~Won Chung, Le~Hou, Shayne Longpre, Barret Zoph, Yi~Tay, William Fedus, Eric Li, Xuezhi Wang, Mostafa Dehghani, Siddhartha Brahma, et~al. 2022.
\newblock Scaling instruction-finetuned language models.
\newblock \emph{arXiv preprint arXiv:2210.11416}.

\bibitem[{Coavoux et~al.(2019)Coavoux, Elsahar, and Gall{\'e}}]{coavoux2019unsupervised}
Maximin Coavoux, Hady Elsahar, and Matthias Gall{\'e}. 2019.
\newblock Unsupervised aspect-based multi-document abstractive summarization.
\newblock In \emph{Proceedings of the 2nd Workshop on New Frontiers in Summarization}, pages 42--47.

\bibitem[{Dettmers et~al.(2023)Dettmers, Pagnoni, Holtzman, and Zettlemoyer}]{dettmers2023qlora}
Tim Dettmers, Artidoro Pagnoni, Ari Holtzman, and Luke Zettlemoyer. 2023.
\newblock Qlora: Efficient finetuning of quantized llms.
\newblock \emph{arXiv preprint arXiv:2305.14314}.

\bibitem[{Erkan and Radev(2004)}]{erkan2004lexrank}
G{\"u}nes Erkan and Dragomir~R Radev. 2004.
\newblock Lexrank: Graph-based lexical centrality as salience in text summarization.
\newblock \emph{Journal of artificial intelligence research}, 22:457--479.

\bibitem[{Gao et~al.(2024)Gao, Hu, Ruan, Pu, and Wan}]{gao2024llm}
Mingqi Gao, Xinyu Hu, Jie Ruan, Xiao Pu, and Xiaojun Wan. 2024.
\newblock Llm-based nlg evaluation: Current status and challenges.
\newblock \emph{arXiv preprint arXiv:2402.01383}.

\bibitem[{Goldsack et~al.(2023)Goldsack, Luo, Xie, Scarton, Shardlow, Ananiadou, and Lin}]{goldsack2023overview}
Tomsa Goldsack, Zheheng Luo, Qianqian Xie, Carolina Scarton, Matthew Shardlow, Sophia Ananiadou, and Chenghua Lin. 2023.
\newblock Overview of the biolaysumm 2023 shared task on lay summarization of biomedical research articles.
\newblock \emph{arXiv preprint arXiv:2309.17332}.

\bibitem[{Guha et~al.(2021)Guha, Mullick, Agrawal, Ram, Ghui, Lee, Bhattacharjee, and Goyal}]{guha2021matscie}
Souradip Guha, Ankan Mullick, Jatin Agrawal, Swetarekha Ram, Samir Ghui, Seung-Cheol Lee, Satadeep Bhattacharjee, and Pawan Goyal. 2021.
\newblock Matscie: An automated tool for the generation of databases of methods and parameters used in the computational materials science literature.
\newblock \emph{Computational Materials Science (Comput. Mater. Sci.)}, 192:110325.

\bibitem[{Hayashi et~al.(2021)Hayashi, Budania, Wang, Ackerson, Neervannan, and Neubig}]{hayashi2021wikiasp}
Hiroaki Hayashi, Prashant Budania, Peng Wang, Chris Ackerson, Raj Neervannan, and Graham Neubig. 2021.
\newblock Wikiasp: A dataset for multi-domain aspect-based summarization.
\newblock \emph{Transactions of the Association for Computational Linguistics}, 9:211--225.

\bibitem[{Kieser et~al.(2023)Kieser, Wulff, Kuhn, and K{\"u}chemann}]{kieser2023educational}
Fabian Kieser, Peter Wulff, Jochen Kuhn, and Stefan K{\"u}chemann. 2023.
\newblock Educational data augmentation in physics education research using chatgpt.
\newblock \emph{Physical Review Physics Education Research}, 19(2):020150.

\bibitem[{Lewis et~al.(2019)Lewis, Liu, Goyal, Ghazvininejad, Mohamed, Levy, Stoyanov, and Zettlemoyer}]{lewis2019bart}
Mike Lewis, Yinhan Liu, Naman Goyal, Marjan Ghazvininejad, Abdelrahman Mohamed, Omer Levy, Ves Stoyanov, and Luke Zettlemoyer. 2019.
\newblock Bart: Denoising sequence-to-sequence pre-training for natural language generation, translation, and comprehension.
\newblock \emph{arXiv preprint arXiv:1910.13461}.

\bibitem[{Li et~al.(2023)Li, Li, Liu, Xie, Li, Wang, Li, and Zhong}]{li2023label}
Zongxi Li, Xianming Li, Yuzhang Liu, Haoran Xie, Jing Li, Fu-lee Wang, Qing Li, and Xiaoqin Zhong. 2023.
\newblock Label supervised llama finetuning.
\newblock \emph{arXiv preprint arXiv:2310.01208}.

\bibitem[{Lin(2004)}]{lin2004rouge}
Chin-Yew Lin. 2004.
\newblock Rouge: A package for automatic evaluation of summaries.
\newblock In \emph{Text summarization branches out}, pages 74--81.

\bibitem[{Liu et~al.(2023)Liu, Yang, Huang, Zhang, Huang, Wei, Deng, Sun, and Zhang}]{liu2023calibrating}
Yuxuan Liu, Tianchi Yang, Shaohan Huang, Zihan Zhang, Haizhen Huang, Furu Wei, Weiwei Deng, Feng Sun, and Qi~Zhang. 2023.
\newblock Calibrating llm-based evaluator.
\newblock \emph{arXiv preprint arXiv:2309.13308}.

\bibitem[{Luo et~al.(2022)Luo, Xie, and Ananiadou}]{luo2022readability}
Zheheng Luo, Qianqian Xie, and Sophia Ananiadou. 2022.
\newblock Readability controllable biomedical document summarization.
\newblock In \emph{Findings of the Association for Computational Linguistics: EMNLP 2022}, pages 4667--4680.

\bibitem[{Maddigan and Susnjak(2023)}]{maddigan2023chat2vis}
Paula Maddigan and Teo Susnjak. 2023.
\newblock Chat2vis: Generating data visualisations via natural language using chatgpt, codex and gpt-3 large language models.
\newblock \emph{IEEE Access}.

\bibitem[{Meng et~al.(2022)Meng, Huang, Zhang, and Han}]{meng2022generating}
Yu~Meng, Jiaxin Huang, Yu~Zhang, and Jiawei Han. 2022.
\newblock Generating training data with language models: Towards zero-shot language understanding.
\newblock \emph{Advances in Neural Information Processing Systems}, 35:462--477.

\bibitem[{Mihalcea and Tarau(2004)}]{mihalcea2004textrank}
Rada Mihalcea and Paul Tarau. 2004.
\newblock Textrank: Bringing order into text.
\newblock In \emph{Proceedings of the 2004 conference on empirical methods in natural language processing}, pages 404--411.

\bibitem[{Mukherjee et~al.(2020)Mukherjee, Peruri, Vishnu, Goyal, Bhattacharya, and Ganguly}]{mukherjee2020read}
Rajdeep Mukherjee, Hari~Chandana Peruri, Uppada Vishnu, Pawan Goyal, Sourangshu Bhattacharya, and Niloy Ganguly. 2020.
\newblock Read what you need: Controllable aspect-based opinion summarization of tourist reviews.
\newblock In \emph{Proceedings of the 43rd international ACM SIGIR conference on research and development in information retrieval}, pages 1825--1828.

\bibitem[{Mullick(2023{\natexlab{a}})}]{mullickexploring}
Ankan Mullick. 2023{\natexlab{a}}.
\newblock Exploring multilingual intent dynamics and applications.
\newblock \emph{IJCAI Doctoral Consortium}.

\bibitem[{Mullick(2023{\natexlab{b}})}]{mullick2023novel}
Ankan Mullick. 2023{\natexlab{b}}.
\newblock Novel intent detection and active learning based classification (student abstract).
\newblock \emph{arXiv e-prints}, pages arXiv--2304.

\bibitem[{Mullick et~al.(2024)Mullick, Ghosh, Chaitanya, Ghui, Nayak, Lee, Bhattacharjee, and Goyal}]{mullick2024matscire}
Ankan Mullick, Akash Ghosh, G~Sai Chaitanya, Samir Ghui, Tapas Nayak, Seung-Cheol Lee, Satadeep Bhattacharjee, and Pawan Goyal. 2024.
\newblock Matscire: Leveraging pointer networks to automate entity and relation extraction for material science knowledge-base construction.
\newblock \emph{Computational Materials Science}, 233:112659.

\bibitem[{Mullick et~al.(2018{\natexlab{a}})Mullick, Ghosh~D, Maheswari, Sahoo, Maity, and Goyal}]{mullick2018identifying}
Ankan Mullick, Surjodoy Ghosh~D, Shivam Maheswari, Srotaswini Sahoo, Suman~Kalyan Maity, and Pawan Goyal. 2018{\natexlab{a}}.
\newblock Identifying opinion and fact subcategories from the social web.
\newblock In \emph{Proceedings of the 2018 ACM International Conference on Supporting Group Work}, pages 145--149.

\bibitem[{Mullick et~al.(2016)Mullick, Goyal, and Ganguly}]{mullick2016graphical}
Ankan Mullick, Pawan Goyal, and Niloy Ganguly. 2016.
\newblock A graphical framework to detect and categorize diverse opinions from online news.
\newblock In \emph{Proceedings of the Workshop on Computational Modeling of People’s Opinions, Personality, and Emotions in Social Media (PEOPLES)}, pages 40--49.

\bibitem[{Mullick et~al.(2017{\natexlab{a}})Mullick, Goyal, Ganguly, and Gupta}]{mullick2017extracting}
Ankan Mullick, Pawan Goyal, Niloy Ganguly, and Manish Gupta. 2017{\natexlab{a}}.
\newblock Extracting social lists from twitter.
\newblock In \emph{Proceedings of the 2017 IEEE/ACM International Conference on Advances in Social Networks Analysis and Mining 2017}, pages 391--394.

\bibitem[{Mullick et~al.(2018{\natexlab{b}})Mullick, Goyal, Ganguly, and Gupta}]{mullick2018harnessing}
Ankan Mullick, Pawan Goyal, Niloy Ganguly, and Manish Gupta. 2018{\natexlab{b}}.
\newblock Harnessing twitter for answering opinion list queries.
\newblock \emph{IEEE Transactions on Computational Social Systems}, 5(4):1083--1095.

\bibitem[{Mullick et~al.(2017{\natexlab{b}})Mullick, Maheshwari, Goyal, and Ganguly}]{mullick2017generic}
Ankan Mullick, Shivam Maheshwari, Pawan Goyal, and Niloy Ganguly. 2017{\natexlab{b}}.
\newblock A generic opinion-fact classifier with application in understanding opinionatedness in various news section.
\newblock In \emph{Proceedings of the 26th International Conference on World Wide Web Companion}, pages 827--828.

\bibitem[{Mullick et~al.(2023)Mullick, Mondal, Ray, Raghav, Chaitanya, and Goyal}]{mullick2023intent}
Ankan Mullick, Ishani Mondal, Sourjyadip Ray, R~Raghav, G~Chaitanya, and Pawan Goyal. 2023.
\newblock Intent identification and entity extraction for healthcare queries in indic languages.
\newblock In \emph{Findings of the Association for Computational Linguistics: EACL 2023}, pages 1825--1836.

\bibitem[{Mullick et~al.(2022{\natexlab{a}})Mullick, Nandy, Kapadnis, Patnaik, Raghav, and Kar}]{mullick2022evaluation}
Ankan Mullick, Abhilash Nandy, Manav Kapadnis, Sohan Patnaik, R~Raghav, and Roshni Kar. 2022{\natexlab{a}}.
\newblock An evaluation framework for legal document summarization.
\newblock In \emph{Proceedings of the Thirteenth Language Resources and Evaluation Conference}, pages 4747--4753.

\bibitem[{Mullick et~al.(2022{\natexlab{b}})Mullick, Nandy, Kapadnis, Patnaik, and Raghav}]{mullick2022fine}
Ankan Mullick, Abhilash Nandy, Manav~Nitin Kapadnis, Sohan Patnaik, and R~Raghav. 2022{\natexlab{b}}.
\newblock Fine-grained intent classification in the legal domain.
\newblock \emph{arXiv preprint arXiv:2205.03509}.

\bibitem[{Mullick et~al.(2022{\natexlab{c}})Mullick, Pal, Nayak, Lee, Bhattacharjee, and Goyal}]{mullick2022using}
Ankan Mullick, Shubhraneel Pal, Tapas Nayak, Seung-Cheol Lee, Satadeep Bhattacharjee, and Pawan Goyal. 2022{\natexlab{c}}.
\newblock Using sentence-level classification helps entity extraction from material science literature.
\newblock In \emph{Proceedings of the Thirteenth Language Resources and Evaluation Conference}, pages 4540--4545.

\bibitem[{Mullick et~al.(2019)Mullick, Pal, Chanda, Panigrahy, Bharadwaj, Singh, and Dam}]{mullick2019d}
Ankan Mullick, Sourav Pal, Projjal Chanda, Arijit Panigrahy, Anurag Bharadwaj, Siddhant Singh, and Tanmoy Dam. 2019.
\newblock D-fj: Deep neural network based factuality judgment.
\newblock \emph{Technology}, 50:173.

\bibitem[{Mullick et~al.(2022{\natexlab{d}})Mullick, Purkayastha, Goyal, and Ganguly}]{mullick2022framework}
Ankan Mullick, Sukannya Purkayastha, Pawan Goyal, and Niloy Ganguly. 2022{\natexlab{d}}.
\newblock A framework to generate high-quality datapoints for multiple novel intent detection.
\newblock \emph{arXiv preprint arXiv:2205.02005}.

\bibitem[{Papineni et~al.(2002)Papineni, Roukos, Ward, and Zhu}]{papineni2002bleu}
Kishore Papineni, Salim Roukos, Todd Ward, and Wei-Jing Zhu. 2002.
\newblock Bleu: a method for automatic evaluation of machine translation.
\newblock In \emph{Proceedings of the 40th annual meeting of the Association for Computational Linguistics}, pages 311--318.

\bibitem[{Paulus et~al.(2017)Paulus, Xiong, and Socher}]{paulus2017deep}
Romain Paulus, Caiming Xiong, and Richard Socher. 2017.
\newblock A deep reinforced model for abstractive summarization.
\newblock \emph{arXiv preprint arXiv:1705.04304}.

\bibitem[{Penedo et~al.(2023)Penedo, Malartic, Hesslow, Cojocaru, Cappelli, Alobeidli, Pannier, Almazrouei, and Launay}]{penedo2023refinedweb}
Guilherme Penedo, Quentin Malartic, Daniel Hesslow, Ruxandra Cojocaru, Alessandro Cappelli, Hamza Alobeidli, Baptiste Pannier, Ebtesam Almazrouei, and Julien Launay. 2023.
\newblock The refinedweb dataset for falcon llm: outperforming curated corpora with web data, and web data only.
\newblock \emph{arXiv preprint arXiv:2306.01116}.

\bibitem[{Raffel et~al.(2020)Raffel, Shazeer, Roberts, Lee, Narang, Matena, Zhou, Li, and Liu}]{raffel2020exploring}
Colin Raffel, Noam Shazeer, Adam Roberts, Katherine Lee, Sharan Narang, Michael Matena, Yanqi Zhou, Wei Li, and Peter~J Liu. 2020.
\newblock Exploring the limits of transfer learning with a unified text-to-text transformer.
\newblock \emph{The Journal of Machine Learning Research}, 21(1):5485--5551.

\bibitem[{See et~al.(2017)See, Liu, and Manning}]{see2017get}
Abigail See, Peter~J Liu, and Christopher~D Manning. 2017.
\newblock Get to the point: Summarization with pointer-generator networks.
\newblock \emph{arXiv preprint arXiv:1704.04368}.

\bibitem[{Sun et~al.(2023)Sun, Ong, Kennedy, Tang, Chen, Elias, Lucas, Shih, and Peng}]{sun2023evaluating}
Zhaoyi Sun, Hanley Ong, Patrick Kennedy, Liyan Tang, Shirley Chen, Jonathan Elias, Eugene Lucas, George Shih, and Yifan Peng. 2023.
\newblock Evaluating gpt-4 on impressions generation in radiology reports.
\newblock \emph{Radiology}, 307(5):e231259.

\bibitem[{Vig et~al.(2021)Vig, Fabbri, Kry{\'s}ci{\'n}ski, Wu, and Liu}]{vig2021exploring}
Jesse Vig, Alexander~R Fabbri, Wojciech Kry{\'s}ci{\'n}ski, Chien-Sheng Wu, and Wenhao Liu. 2021.
\newblock Exploring neural models for query-focused summarization.
\newblock \emph{arXiv preprint arXiv:2112.07637}.

\bibitem[{Xu and Lapata(2020)}]{xu2020query}
Yumo Xu and Mirella Lapata. 2020.
\newblock Query focused multi-document summarization with distant supervision.
\newblock \emph{arXiv preprint arXiv:2004.03027}.

\bibitem[{Yang et~al.(2022)Yang, Song, Cho, Wang, Pan, Petzold, and Yu}]{yang2022oasum}
Xianjun Yang, Kaiqiang Song, Sangwoo Cho, Xiaoyang Wang, Xiaoman Pan, Linda Petzold, and Dong Yu. 2022.
\newblock Oasum: Large-scale open domain aspect-based summarization.
\newblock \emph{arXiv preprint arXiv:2212.09233}.

\bibitem[{Ye et~al.(2022)Ye, Gao, Feng, Wu, Yu, and Kong}]{ye2022progen}
Jiacheng Ye, Jiahui Gao, Jiangtao Feng, Zhiyong Wu, Tao Yu, and Lingpeng Kong. 2022.
\newblock Progen: Progressive zero-shot dataset generation via in-context feedback.
\newblock \emph{arXiv preprint arXiv:2210.12329}.

\bibitem[{Yu et~al.(2023)Yu, Zhuang, Zhang, Meng, Ratner, Krishna, Shen, and Zhang}]{yu2023large}
Yue Yu, Yuchen Zhuang, Jieyu Zhang, Yu~Meng, Alexander Ratner, Ranjay Krishna, Jiaming Shen, and Chao Zhang. 2023.
\newblock Large language model as attributed training data generator: A tale of diversity and bias.
\newblock \emph{arXiv preprint arXiv:2306.15895}.

\bibitem[{Zhang et~al.(2020)Zhang, Zhao, Saleh, and Liu}]{zhang2020pegasus}
Jingqing Zhang, Yao Zhao, Mohammad Saleh, and Peter Liu. 2020.
\newblock Pegasus: Pre-training with extracted gap-sentences for abstractive summarization.
\newblock In \emph{International Conference on Machine Learning}, pages 11328--11339. PMLR.

\bibitem[{Zhang et~al.(2019)Zhang, Kishore, Wu, Weinberger, and Artzi}]{zhang2019bertscore}
Tianyi Zhang, Varsha Kishore, Felix Wu, Kilian~Q Weinberger, and Yoav Artzi. 2019.
\newblock Bertscore: Evaluating text generation with bert.
\newblock \emph{arXiv preprint arXiv:1904.09675}.

\bibitem[{Zhou et~al.(2023)Zhou, Zhu, Chen, Chen, Zhao, Chen, Lin, Wen, and Han}]{zhou2023don}
Kun Zhou, Yutao Zhu, Zhipeng Chen, Wentong Chen, Wayne~Xin Zhao, Xu~Chen, Yankai Lin, Ji-Rong Wen, and Jiawei Han. 2023.
\newblock Don't make your llm an evaluation benchmark cheater.
\newblock \emph{arXiv preprint arXiv:2311.01964}.

\bibitem[{Zhu et~al.(2022)Zhu, Dong, Wei, Qin, and Liu}]{zhu2022transforming}
Haichao Zhu, Li~Dong, Furu Wei, Bing Qin, and Ting Liu. 2022.
\newblock Transforming wikipedia into augmented data for query-focused summarization.
\newblock \emph{IEEE/ACM Transactions on Audio, Speech, and Language Processing}, 30:2357--2367.

\end{thebibliography}

\newpage

\appendix

\section*{Appendix}

\section{Related Work}
In this section, we conduct a survey of state-of-the-art literature on closely related topics.

\noindent \textbf{Summarization:} We perform a survey of various summarization techniques, that encompasses generic approaches such as abstractive~\cite{chopra2016abstractive, see2017get, paulus2017deep} and extractive summarization like TextRank \cite{mihalcea2004textrank} and LexRank \cite{erkan2004lexrank}. Subsequent research has explored summarization in specific directions, including aspect-based summaries~\cite{hayashi2021wikiasp, coavoux2019unsupervised, mukherjee2020read, akhtar2017aspect} and query-focused summaries~\cite{vig2021exploring, zhu2022transforming, xu2020query}, but none focuses on domain-specific persona-based summarization.

\noindent \textbf{Persona Concept:} \cite{goldsack2023overview, luo2022readability} focus on building lay-summarization for comprehensible to non-technical audiences but do not have the differentiating factor of persona concept to distinguish the various technical summaries. based on persona and no usage of LLMs as an alternative metrics to evaluate the summaries. Our work differs from the fact that we show a pipelined approach - data generation, persona-specific (doctor vs patient vs normal person) summarization with key points and GPT-4 based evaluation to save time and assist different persona to augment their knowledge on it to make conclusions more efficiently. The concept of Persona also different from the notion of intent~\cite{mullick2022framework, mullick2023intent, mullick2022fine, mullick2023novel, mullickexploring, mullick2022evaluation}, entity-relation~\cite{mullick2024matscire,guha2021matscie,mullick2022using} or opinion/fact~\cite{mullick2017generic, mullick2016graphical, mullick2018identifying, mullick2018harnessing, mullick2019d,mullick2017extracting} idea. Our work differs from the fact that we show a pipelined approach - data generation, `persona concept' specific (doctor vs patient vs normal person) summarization with key points and GPT-4 based evaluation to save time and assist different persona to augment their knowledge on it to make conclusions more efficiently.

\noindent \textbf{Data Generation using LLMs:} Large Language Models (LLMs), such as the GPT family, have been utilized to generate training datasets for NLP tasks, addressing data scarcity issues in a cost effective manner~\cite{yu2023large, meng2022generating, ye2022progen}. ChatGPT~\footnote{https://chat.openai.com/} aids in educational data augmentation and data visualization~\cite{kieser2023educational, maddigan2023chat2vis}; however, in healthcare domain, GPT-3 and GPT-4 are employed for generating medical dialogue summarization data~\cite{chintagunta2021medically} and radiology reports~\cite{sun2023evaluating}. However, no prior work focuses on benchmark domain-specific persona-based summary data generation with appropriate human validation.

\noindent \textbf{LLMs as Evaluation Metrics:} The rise of LLMs presents a potential, cost-effective alternative for evaluating various NLP tasks. Existing efforts include a taxonomy of LLM-based NLG evaluation methods~\cite{gao2024llm}, the development of `ChatEval' for assessing response quality~\cite{chan2023chateval}, and proposed guidelines for LLM-based evaluation~\cite{zhou2023don}. While some studies explore LLM-based assessments with human alignment~\cite{liu2023calibrating}, there is a lack of work utilizing LLMs to evaluate solution architectures comprehensively with a critique scoring system.

Our paper takes a significant step towards addressing the shortcomings of prior literature in terms of persona-based summary data generation followed by human valiadation as well as performing LLM based critic evaluation.

\label{appendix:related_works}

\section{Dataset Requirement} \label{appendix : dataset_requirement}

Access to robust, comprehensive datasets is crucial for training NLP models to understand and generate persona-specific content, emphasizing the challenges of data scarcity and summary-making capabilities. Manual annotation of large healthcare datasets is both costly and time-intensive, demanding domain expertise and meticulous attention to detail. Consider the example where $n$ individuals generate summaries for $m$ persona, resulting in $n \times m$ distinct summary generations for a single article — it is a highly expensive and time consuming process. Moreover, acquiring suitable human resources for labeling is challenging, as people are often hesitant to undertake the tedious and difficult task of summary generation, even with a standard payment agreement. \\

We provide healthcare data to the Prolific annotation platform with specific criteria: annotators with a PhD / Graduate degree, a medical background, approval rates of 90\%-100\%, and expertise in medicine. However, out of 157,341 potential annotators, only 189 meet these criteria, and even among them, there is a high rejection rate of 71.43\% for selecting documents for manual summary generation, despite offering more than standard payment. Further, eligible annotators tend to heavily rely on ChatGPT and similar automated approaches, leading to the need for extensive re-evaluations and revisions, along with issues related to annotator rejection, even to assess a limited number of documents. Hence, obtaining a high quality dataset remains a formidable challenge.

\section{GPT-4 Bias}

It is acknowledged that using summaries generated exclusively by GPT-4 could introduce biases inherent in its summarization capabilities, it may also be noted that alternatives, such as human evaluation, also carry their own biases. Despite the potential for bias, leveraging GPT-4 for summarization may still be a pragmatic choice, especially in scenarios where access to diverse datasets or sophisticated validation methods is limited. However, in this work, we remain vigilant, recognizing the limitations inherent in both automated and human-generated summaries, and take proactive steps such as human intervention to validate and contextualise the results to mitigate biases to the best extent possible within the given constraints.

\section{Prompts} \label{appendix : prompts}

We use prompting in three stages - data generation, finetune-inference and critique. There are two kinds of prompts - system prompt and user prompt. 

\subsection{Data Generation}

\textbf{system} : You are an AI assistant who are to generate a summary of a medical document specific to a certain persona which can be doctor, patient, normal person. The summary of a medical document should be generated from the perspective of the respective persona. \\
\textbf{user} : Summarize the medical document given below from the perspective of a \{persona\} [doctor/patient/normal person] and return the summary only. The medical document is as follows: Document: \{document\}

\subsection{Finetune and Inference prompt}

\textbf{user prompt} - Summarize the medical document given below from the perspective of a {persona}: \\ 
$\#\#\#$ Document: {document}

\subsection{Critique}

\textbf{system}: You are an AI assistant who is to evaluate the summary of a medical document specific to a certain persona which can be doctor, patient or a normal person. A doctor  requires a detailed and technical summary about the medical document. Patients require a layman's summary about the medical document, with information about  things like causes, effects, treatment etc. that may be helpful to them. A normal person has no medical knowledge and requires a generic summary about the medical document. You need to return a score between 0 and 1 reflecting the quality of the generated summary based on some criteria.\\
\textbf{user}: You are given a medical document and the corresponding summary of the document generated from the perspective of a \{persona\} predicted by a language model as follows. \\
Document: \{document\} \\
Ground truth summary : \{label summary\} \\
Summary from the perspective of a \{persona\} [doctor/patient/normal person]: \{model generated summary\} \\
Evaluate the above persona based summary for the document in terms of each of the following criteria and return only a score between 0 and 1 without any explanation:
\begin{itemize}
    \item The extent to which the generated summary is relevant to the specific persona \{persona\}[doctor/patient/normal person] based summary of the document.
    \item The extent to which the generated persona-based summary correctly covers all the important key points described in the persona \{persona\}[doctor/patient/normal person] based summary of the document.  
    \item The extent to which the summary does not contain information specific to all other possible personas \{persona\_set - persona\}[doctor/patient/normal person] based summary. 
    \item Rate the summary from the point of view of the persona – whether the summary is good, average, or bad. A good summary effectively captures the essential points, presenting them clearly and concisely. It maintains accuracy, encourages reader engagement, and serves as a compelling introduction to the content. An average summary conveys the main points but may lack some clarity or detail, presenting a decent overview without standing out in terms of conciseness or precision. It provides a basic understanding but not from a more refined focused summary and fails to accurately convey the main points, containing inaccuracies or misinterpretations. It is either overly verbose or lacks coherence, making it difficult for the reader to grasp the core information effectively. 
    \item Calculated summary from the point of view of the persona [Good/Bad/Average] [Calculated from 4 with the help of manual annotation]
\end{itemize}

\section{Experiments} \label{appendix: experiments}

\subsection{Varying Training Size Dataset}
\begin{figure}[!htp]
    \centering
    \vspace{-3mm}
    \begin{adjustbox}{width=0.650\linewidth}
\includegraphics[width=\linewidth]{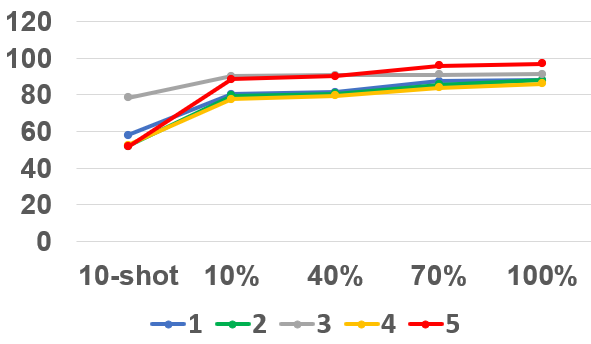}
    \end{adjustbox}
    \vspace{-3mm}
    \caption{Different Training Data Sizes}
    \label{fig:vary-training-only}
    \vspace{-3mm}
\end{figure}

To understand the effect of training data size on the performance, we vary the WebMD training data for the Llama2-13b model - taking  $10$-shot ($k$-shot settings where $k$=$10$), 10\%, 40\% and 70\% of the initial training data, and finetune the Llama2-13b model with same parameter and hyper-parameter settings and the five criterias of GPT-4 Critique outcome (in \%) are shown in Fig \ref{fig:vary-training-only}. We see that with increasing the dataset size, the performance of Llama2-13b improves in terms of GPT-4 critique and traditional metrics. Even at 40\% of the dataset, the model is able to achieve a very good performance. It shows the effectiveness of the Llama2-13b model. It also infers that even with very little amount of data in $10$-shot Llama2-13b can able to generate appropriate persona and aspect based summary.


\subsection{Tuning generation parameters during model inference:} 
\begin{figure}[h]
\centering
\vspace{-3mm}
        \begin{subfigure}[b]{0.25\textwidth}
                \centering \includegraphics[width=0.99\linewidth]{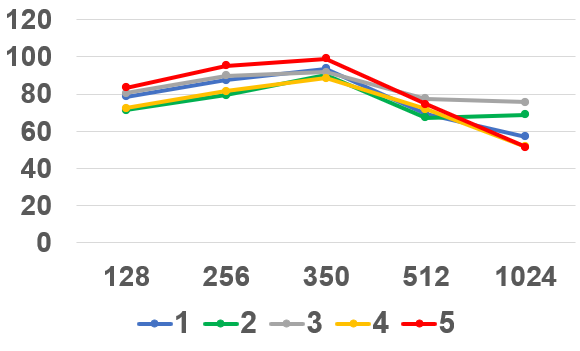}
                \caption{w.r.t. max-new-token}
                \label{fig:max-new-token}
        \end{subfigure}%
        \begin{subfigure}[b]{0.25\textwidth}
                \centering
    \includegraphics[width=0.96\linewidth]{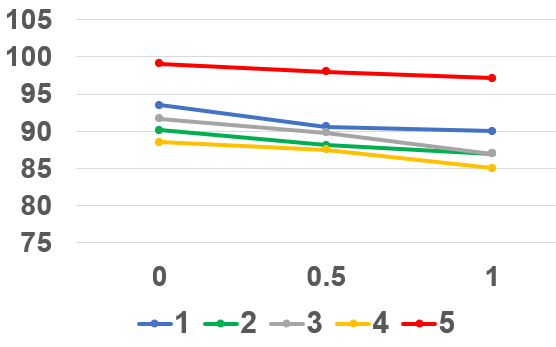}
                \caption{w.r.t. temperature}
                \label{fig:temp}
        \end{subfigure}
        \vspace{-4mm}
        \caption{Variations of Llama2-13b-Finetune (in \%)}
        \vspace{-2mm}
        \label{fig:token-temp}
\end{figure}

We investigate the impact of tuning the \textit{max-new-token} and \textit{temperature} generation parameters on the performance of our finetuned model during inference. 
The variation in performance in terms of the five GPT-4 critique based criteria are shown in Figures~\ref{fig:max-new-token} and~\ref{fig:temp} respectively. We observe that our model performs the best for a temperature of $0$ and performance degrades significantly as we increase the temperature beyond $0.4$. Similarly, the best model performance is achieved for a \textit{max-new-token} size of $350$. We have used NLTK, Spacy, openai (version=0.28), huggingface\_hub, torch and transformers python packages for all experiment. 

\subsection{Time and GPU} \label{time-gpu}
We experiment on 80GB A100 GPU with GPU clock cycle 210 MHz. The finetuning and inference time of our finetuned models are in Table \ref{tab:time-gpu}.

\begin{table}[!ht]
    \centering
    \begin{tabular}{|c|c|c|}\hline
         Model & Finetune Time & Inference Time\\
         \hline
         Llama2-7b & 8 hrs & 2hrs 30 mins \\\hline
         Llama2-13b & 20 hrs & 2hrs 50 mins\\\hline
    \end{tabular}
    \caption{Model Training Time [using 80GB A100 GPU]}
    \label{tab:time-gpu}
\end{table}

\section{Human Annotations} \label{appendix : human_annotations}


\noindent \textbf{Annotation Guidelines for Comparative Rating:}
We provide instructions with explanations of different \textit{persona} and ask to identify which summary belongs to which persona as shown in Fig \ref{fig:exp_1_set_1_exp}. We also provide the link of document along with distinct summaries of GPT-4 and Llama2-13b finetune model for comprison as shown in Fig \ref{fig:exp_2_set_1_exp}. The instructions are the following -\\
``You are given a summary of a medical document specific to the perspective of a certain group of people (doctor, patient and normal person). \\
A doctor requires a detailed and technical summary about the medical document. \\
A patient requires a layman summary about the medical document, with information about  things like causes, effects, treatment etc. that may be helpful to him. A patient only requires a top level view of the extensive medical details and not so much medical details like a doctor.\\
A normal person has no medical knowledge and requires a generic summary about the medical document and does not require extensive medical details.''

\noindent \textbf{Annotation Guidelines to Prolific and Doctors}

For annotator selection, we have several criterias. Annotator selection includes specific criteria such as `Degree subject' in Health and welfare, `Highest education level completed' as Doctorate degree or Graduate degree, `Fluent languages' in English, `Approval Rate' of 90–100, `Subject' in Medicine, and `Employment Sector' as Doctor. Further annotations are conducted by graduate doctors (details in Section - Human Evaluation Part).

Following is the annotation guideline to `Prolific' annotation platform -

\textit{Objective}: Generate Personified Summaries by Prolific

\textit{Introduction}: In this study, you are tasked with generating a summary tailored to three different personas: a Doctor, a Patient, and a Normal Person. You will be provided with a document link containing a Source Document (SD) – which can be a medical research document or a general article related to health from https://www.webmd.com/. . Additionally, a Persona (P) will be present.

\textit{Your Task}: Read the Source Document (and Persona) and craft three summaries, each targeted towards one of the personas mentioned. Use your understanding and perspective to tailor the information in a way that is most relevant and comprehensible to each persona.

\textit{Summary Persona}:

Doctor Persona: Craft a summary that focuses on medical terminology, guidelines, and provides information suitable for a medical professional. Emphasize technical accuracy and relevance to medical practice. A doctor requires a detailed and technical summary about the medical document.

Patient Persona: Generate a summary with a patient-centric approach, avoiding excessive technical jargon. Ensure that the information is clear, easily understandable, and addresses concerns that a patient might have. A patient requires a non-technical summary about the medical document, with information about  things like causes, effects, treatment etc. that may be helpful to him. A patient only requires a top level view of the extensive medical details and not so much medical details like a doctor.

Normal Person Persona: Tailor a summary for a general audience without a medical background. Use simple language, avoid technical terms, and present the information in a way that is accessible and engaging to a layperson. A normal person has no medical knowledge and requires a generic summary about the medical document and does not require extensive medical details.

\textit{Instructions}: 
1. Carefully review the Source Document and the Persona. Consider the specific needs and understanding level of each persona while generating the summaries.

2. No additional software download is required. Use a browser, preferably Google Chrome, and ensure a stable internet connection.

3. Allocate time judiciously for crafting each of the three summaries based on the provided 2 SD instances.

4. After completion, you will be asked to provide feedback on the generation exercise, platform interaction, and details about your academic background, age, country of birth, and any medical background or experience with model-generated summaries.

\textit{Payment Requirements}: Upon completing the study, click on the provided link containing the completion code to redirect you to the Prolific platform. Payment will be processed within one to two weeks.

\textit{Ethical Considerations}:

Adhere to strict confidentiality and data protection standards to ensure the privacy of medical information. If you have concerns or questions, feel free to reach out, as this study aligns with ethical guidelines.

This study aims to harness diverse perspectives, including those of medical professionals, to refine the generation of personified summaries for enhanced utility in various contexts.

Next, the details while providing the documents - 
``You will be given 2 documents in the next 2 pages and you need to write the summaries with respect to Doctor, Patient and Normal Person (as in example). 

Please do not use ChatGPT/GPT-4 or any Large Language Models - all summaries should be generated by human properly. It is a strict instruction and will be checked manually - if found any issue: it will be rejected and re-doing will be required.

Your summary length (word count) is approximately 15\% - 20\% of the document length (word count) for three different types.''

\section{Examples}  \label{appendix: examples}

\begin{figure*}[!ht]
    \centering
    \includegraphics{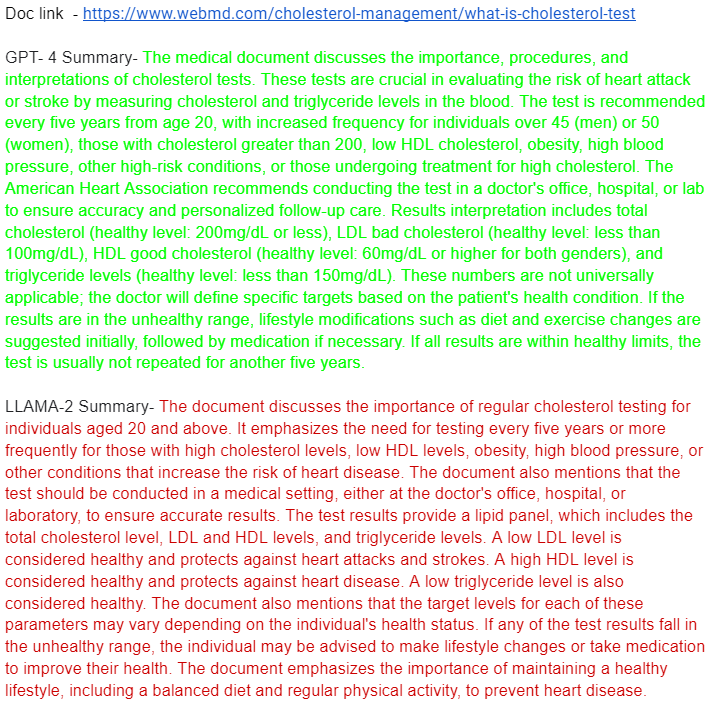}
    \caption{GPT-4 generated summary better than LLAMA2-13b model generated summary[persona : doctor]}
    \label{fig:gpt_better_exp_2_set_1}
\end{figure*}

\begin{figure*}[!ht]
    \centering
    \includegraphics{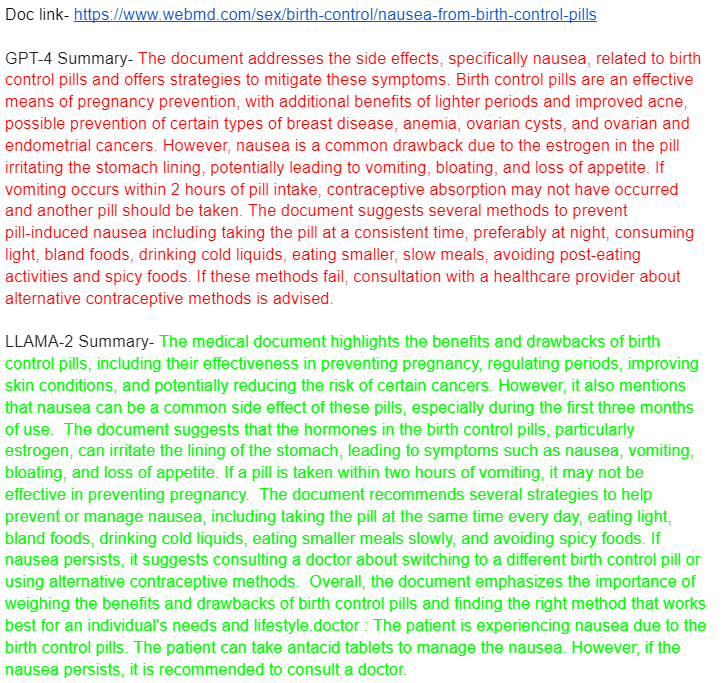}
    \caption{LLAMA2-13b model generated summary better than GPT-4 generated summary[persona : doctor]}
    \label{fig:llama_better_exp_2_set_1}
\end{figure*}

\begin{figure*}[!ht]
    \centering
    \includegraphics{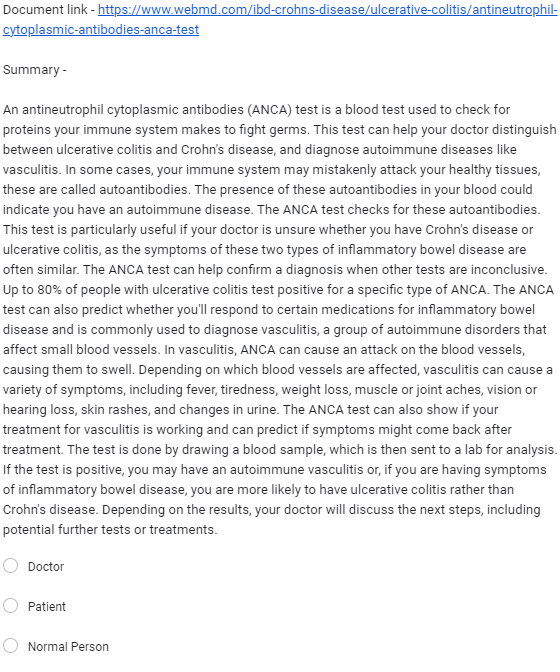}
    \caption{Persona identify experiment example snapshot}
    \label{fig:exp_1_set_1_exp}
\end{figure*}

\begin{figure*}[!ht]
    \centering
    \includegraphics{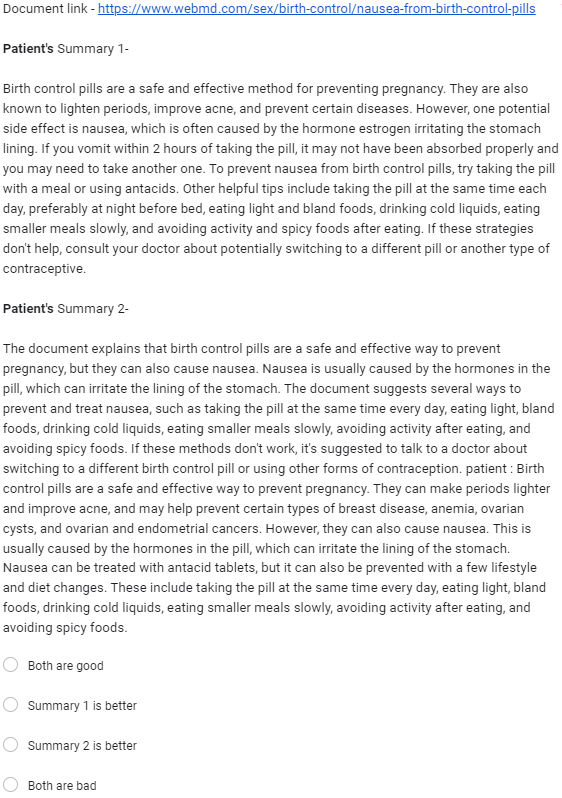}
    \caption{Llama2-13b finetune and GPT-4 summary comparison experiment example snapshot}
    \label{fig:exp_2_set_1_exp}
\end{figure*}

Two human annotated examples in doctor persona is shown in Fig \ref{fig:gpt_better_exp_2_set_1} where the GPT-4 generated summary is better and Fig \ref{fig:llama_better_exp_2_set_1} where LLAMA-2 generated summary is better. Two examples of human annotation interface is shown in Fig \ref{fig:exp_1_set_1_exp} and Fig \ref{fig:exp_2_set_1_exp} respectively.


\end{document}